\documentclass[letterpaper, 10 pt, conference]{ieeeconf}  
\usepackage{amsmath}
\usepackage{cite}
\usepackage{amsmath,amssymb,amsfonts}

\usepackage{algorithmic}
\usepackage[ruled,vlined,linesnumbered]{algorithm2e}

\usepackage{graphicx}
\usepackage{textcomp}
\usepackage{tabularx}
\usepackage{xcolor}
\usepackage{url}
\usepackage{booktabs}
\usepackage{subcaption}

\newif\ifrevisions
\revisionsfalse   

\ifrevisions
    \newcommand{\rev}[1]{\textcolor{blue}{#1}}
    \newcommand{\comm}[1]{\textcolor{orange}{#1}}
\else
    \newcommand{\rev}[1]{#1}
    \newcommand{\comm}[1]{#1}
\fi

\IEEEoverridecommandlockouts                              

\title{\LARGE \bf
How do Transformers Perform Low-Level Control In-Context?
}

\author{Ebonye Smith$^1$, Aidan Andrews$^2$, Phoenix Wilson$^1$, Alexis Frias$^3$, Ayush Pandey$^4$, and Gireeja Ranade$^1$
\thanks{*This material is supported by the National Science Foundation Graduate Research Fellowship under Grant No. DGE-2146752, NSF CAREER ECCS2240031, and 2023 CITRIS Institute seed grant. We also thank Dr. Claire J. Tomlin from UC Berkeley for her valuable input on this paper.}
\thanks{$^{1}$Ebonye Smith (corresponding author), Phoenix Wilson, and Gireeja Ranade are with the Department of Electrical Engineering and Computer Sciences, University of California Berkeley, Berkeley, CA 94720, USA,  {\tt\small \{ebonyesmith, phoenixw, gireeja\}@berkeley.edu}}%
\thanks{$^{2}$Aidan Andrews is with the Physics Department, University of Illinois, Urbana-Champaign,             Urbana, IL 61801
        {\tt\small aidansa2@illinois.edu}}%
\thanks{$^{3}$Alexis Frias is with the Department of Electrical and Computer Engineering, University of Southern California,            Los Angeles, CA 90089
        {\tt\small alexisfr@usc.edu}}%
\thanks{$^{4}$Ayush Pandey is with the Department of Electrical Engineering, University of California Merced,            Merced, CA 95343
        {\tt\small ayushpandey@ucmerced.edu}}%
}

\begin{document}

\maketitle
\thispagestyle{empty}
\pagestyle{empty}

\begin{abstract}
One of the remarkable abilities of Transformer-based neural networks is in-context learning. This motivates us to think about ``in-context'' control. Can we control a system knowing only (1) that it belongs to a particular family of systems and (2) some data on system evolution (provided as the context trajectory) but no specific parameter information?  
Specifically, we investigate the ability of Transformer-models to perform ``in-context'' control of unstable, nonlinear dynamical systems, where context consists only of the unforced  system dynamics. We leverage a multi-head, GPT-2-style architecture, and we evaluate performance for two classical nonlinear systems -- the cartpole system and the two-link acrobot system, where the system parameters are unknown to the Transformer. We consider a setting where trajectories from an expert controller are available at training time but not at inference time. Here, we identify training techniques that are essential to train Transformers to perform low-level control. Through an analysis over checkpoints, we show that in-context control is possible, and the learned policy performance generally improves as context increases. Furthermore, the Transformer models can generalize to stabilize systems that are out-of-distribution. 



\end{abstract}

\section{Introduction}
\label{introduction}
\rev{A key property of Transformers driving its performance}
is in-context learning (ICL), which \rev{enables} the model \rev{to adapt to new tasks at inference time via a context prompt}
without updating model weights~\cite{brown2020language, lieber2021jurassic, rae2021scaling, black2022gpt}. 
While ICL has been studied for linear regression~\cite{garg2022can}, filtering~\cite{DuOymakTrans}, and mid-level robotic control~\cite{fu2024context}, its application to the low-level, closed-loop stabilization of unstable, underactuated, nonlinear systems remains largely unexplored. 

Stabilizing an unstable system \rev{with partially unknown dynamics traditionally requires system identification}~\cite{ljung1987theory} 
or direct \rev{synthesis of a data-driven controller}~\cite{hou2013model}. 
Building on the paradigm of universal control~\cite{busetto2025one}, \rev{we aim to realize a single}
policy capable of stabilizing \rev{an entire parametric family of systems without explicit parameter estimation or weight updates.}
\rev{Under this setting,}
parameter identification is shifted from offline pre-processing to an in-context operation at test time.
\begin{figure}[t!]
    \centering
    \includegraphics[width=0.89\linewidth]{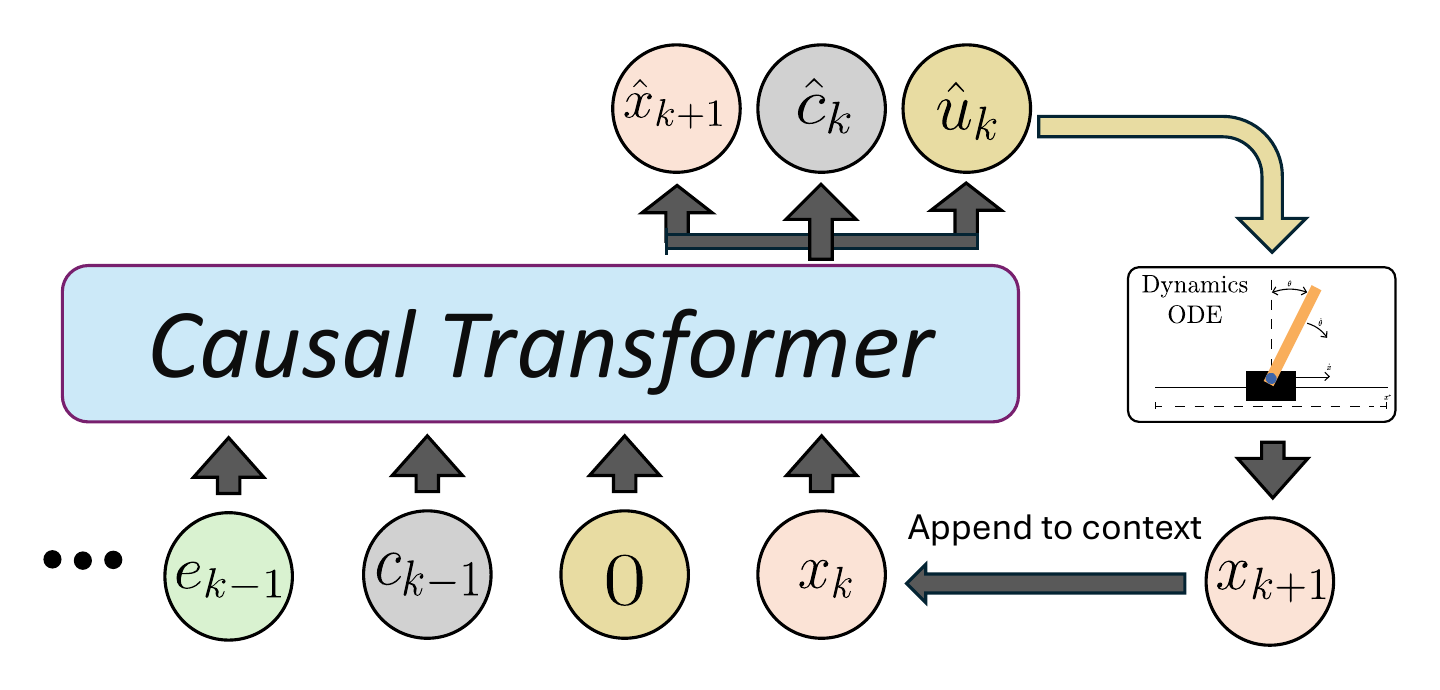}
    \caption{Given a context of zero-torque system evolution \(u_{0:k-1}{=}0\) and current state \(x_k\), the Transformer predicts the next state \(\hat{x}_{k+1}\), controller mode \(\hat{c}_k\), and action \(\hat{u}_k\), which is applied to the plant.}
    \label{fig:inference_fig}
\end{figure}

\comm{Reinforcement learning algorithms are commonly used for  tasks like stabilization of underactuated systems~\cite{deisenroth2011pilco, kober2013reinforcement}}, however we are interested in investigating how ICL capabilities of Transformers along with architectural and training design choices contribute to similar task success.
The primary contributions of this work are\footnote{Our code can be found at the following github repository: \url{https://github.com/ebonyelsmith/transformer_control}}:

\begin{itemize}
    \item \textbf{Analysis of emergent in-context capabilities:} 
    We investigate how stabilization scales with context length and model training \rev{
    demonstrating that ICL emerges as training progresses.}
    Extended training shows evidence of data memorization \rev{which contributes to some cases of successful generalization to out-of-distribution data}. 
    \item \textbf{Robustness via Gaussian perturbations:} We inject state-level Gaussian noise during expert data generation. Without this, in-distribution stabilization success on the Cartpole collapses to 5\% with 50 context examples (from 86\% with perturbation), confirming that exposure to
    driving disturbances proves essential.
    \item \textbf{Empirical validation on nonlinear benchmarks:} We evaluate \rev{our architecture on} underactuated Cartpole~\cite{aastrom2021feedback} and Acrobot~\cite{murray1991case} systems,
    achieving up to 86\% (Cartpole) and 74\% (Acrobot) in-distribution stabilization success at the longest context length.
\end{itemize}

\section{Related Work}
The development of a universal controller for unstable, nonlinear systems sits at the intersection of ICL, sequence-based decision-making, and foundation models for robotics. 

\paragraph{Emergence of In-Context Learning}
The capacity for Transformers to perform ICL
was initially characterized in simple settings, such as linear regression~\cite{garg2022can, raventos2023pretraining} and Gauss-Markov process prediction~\cite{DuOymakTrans}. While 
\rev{extensively studied in large language models}~\cite{brown2020language,olsson2022context,akyurek2024context}, recent \rev{work indicates} ICL is an emergent property driven by the distributional properties of training data~\cite{chan2022data, raparthy2023generalization}, and~\cite{daniels2025different} investigates how ICL relates to dynamical systems prediction and associative recall. 
\rev{We extend these findings to low-level stabilization of unstable, underactuated nonlinear systems.}

\paragraph{Transformers for Sequence-Based Control}
Decision Transformer~\cite{chen2021decision} and Trajectory Transformer~\cite{NEURIPS2021_099fe6b0} catalyzed the use of Transformers within the sequence-modeling approach to reinforcement learning.
Subsequent \rev{works applied these models to system identification~\cite{zhang2025dynamicspromptsincontextlearning} and task adaptation~\cite{xu2022prompting}.}
\rev{Existing approaches focus on high-level planning, however low-level stabilization at unstable equilibria is an interesting case study because of unique sensitivity to compounding errors.}
\rev{We bridge this gap using a multi-task objective for implicit parameter identification and state-perturbed expert demonstrations to ensure stabilization robustness.}

\paragraph{Universal Controllers and Foundation Models}
\rev{We build on universal control paradigms 
\cite{busetto2025one}, which unifies control across plant realizations.}
Parallel efforts in robotics 
\rev{leverage} fine-tuned LLMs~\cite{zhang2025decision, wang2024prompt} or vision-language models~\cite{li2023vision,zhuInCoRoInContextLearning2024a} for closed-loop control. \rev{Unlike foundation models that rely on high-level abstractions, we depart from natural language prompting \cite{wang2024prompt}.}

\section{Problem Formulation}
\label{problem_form}

\subsection{System Representation and Discretization}
\label{system_rep}
Consider a family of nonlinear, continuous-time dynamical systems  \(\mathcal{F}\) governed by the ordinary differential equation
\(
    \dot{x}(t){=}f(x(t),u(t);\omega)
    \)
where \(x(t) \in \mathcal{X}\subseteq \mathbb{R}^n\) is the state, \(u(t) \in \mathcal{U} \subseteq \mathbb{R}^m\) is the control input, \rev{and \(\omega \in \Omega\subset \mathbb{R}^p\) represents unknown physical parameters.}
\rev{Operating at}
sampling interval \(\delta_t\), 
\rev{discrete-time dynamics are given by}
\(
    x_{k+1}{=}\Phi_\omega(x_k, u_k, \delta_t).
\)
We can define the system distribution as
\(
    \mathcal{D}:=\{\Phi_\omega\mid \omega\in \Omega \}.
\)
For each realization \(\omega\), an expert controller \(\kappa_\omega:\mathbb{R}^n \to \mathbb{R}^m\) \rev{yields the control action given the state}
\(x_k\) as \(u_k^\ast{=}\kappa_\omega(x_k)\).
We use \(\kappa_\omega\) for offline training data generation; during inference, the learned policy receives state-action history alone without access to \(\omega\). 

\subsection{Universal Control via In-Context Learning}
\label{univ_ctrl}
Following the idea of universal control~\cite{busetto2025one}, \rev{our goal} is to train a policy \(\pi\) capable of stabilizing any system realization \(\Phi_\omega \sim \mathcal{D}\) without explicit knowledge of \(\omega\), \rev{leveraging ICL via an initial unforced response}.
 Let \(N\) be the window of the unforced response, where we impose zero torque \(u_k {=} 0\). 
The sequence generated during the unforced phase forms the context prompt:
\(
    H_{0:N-1}{=}\{(x_k,e_k,c_k,u_k = 0)\}_{k=0}^{N-1}.
\)
Here, \(e_k \in \mathbb{R}\) is the norm distance between the current state and the goal state, and \(c_k\) represents the active controller mode, \rev{which is necessary in our case as we employ a hybrid expert controller.}
\comm{Because the system model is known, the dynamics can be expressed linearly in the physical parameters $\omega \in \Omega$, and estimating $\omega$ from a trajectory \(H_{0:N-1}\) requires a minimum context window \(N\) to yield a uniquely solvable system of equations~\cite{atkeson1986estimation}.
This motivates our empirical study of stabilization success as a function of \(N\).}

\rev{The policy \(\pi\) uses \(H_{k-N:k-1}\) at step \(k \geq N\) to produce an action approximating the expert:
\(    \pi(H_{k-N:k-1},x_k) \approx \kappa_\omega(x_k)\).}
To optimize \(\pi\), we first construct the training dataset \(\mathcal{T}{=}\{\tau^{(j)}_{\kappa_\omega}\}_{j=1}^M\), where each trajectory \(\tau^{(j)}_{\kappa_\omega}\) is generated \rev{by rolling out trajectories of length \(N_{\text{total}}\) under a sampled realization \(\Phi_\omega \sim \mathcal{D}\).
Each rollout transitions from an \emph{Unforced Response Phase} (\(0 \leq k < N\), where \(u_k{=}0\)) to a \emph{Control Phase} (\(N\leq k \leq N_{\text{total}}\), where \(u_k{=}\kappa_\omega(x_k)\)), stored as \((x_k, e_k, c_k, u_k)\). Rollouts are sliced to length \(L \ll N_{\text{total}}\), with the starting index sampled with equal probability from the unforced response or the full sequence, ensuring frequent exposure to unforced sequences of varying length.}

\section{Multi-task Sequence Model Framework}
\subsection{Transformer Models}
\label{transformer_models}
\rev{The universal policy \(\pi_\gamma\), parameterized by weights \(\gamma \in \Gamma\), is modeled using a GPT-2-style decoder-only Transformer~\cite{radford2018improving}. The model processes context history via causal-self attention masking, only allowing attention
to previous tokens~\cite{j.2018generating,radford2018improving}.
To preserve relative positional awareness across context sequences,
we employ Rotary Position Embeddings (RoPE)~\cite{su2024roformer}.}
\rev{To ground the latent embedding in both system dynamics and control law synthesis, the Transformer branches into three}
Multi-Layer Perceptron (MLP) heads at each step \(k\):
(1) \rev{\textit{Control Head (\(\hat{u}_k\)):} predicts the control action. (2) \textit{State Dynamics Head (\(\hat{x}_{k+1}\)):} predicts the state as an auxiliary task, forcing the shared representation to implicitly identify \(\omega\). (3) \textit{Mode Head (\(\hat{c}_k\)):} A classifier that predicts the active mode of the controller
(e.g., energy-shaping, or stabilization).} 

\subsection{Multi-task Learning Objective}
We obtain \(\pi_\gamma\) by optimizing \(\gamma\) across the system distribution \(\mathcal{D}\) by minimizing a composite loss function \(\mathcal{L}\):
\begin{equation}
\label{eq:loss_eqn}
    \min_\gamma \mathbb{E}_{\Phi_\omega \sim \mathcal{D}}\left[\sum_{k=0}^{T} \mathcal{L}\Big( \pi_\gamma(H_{k-L:k-1}, x_k),{z}_k \Big) \right]
\end{equation}
where \({z}_k{=}(u^\ast_k,x_{k+1},c_k)\)
\rev{contains expert action \(u^\ast_k{=}\kappa_\omega(x_k)\), the next state \(x_{k+1}\), and the controller mode \(c_k\)}, and \(\mathcal{L} {=} \lambda_u \| u_k^\ast - \hat{u}_k \|_2^2 + \lambda_x \| x_{k+1} - \hat{x}_{k+1} \|_2^2 + \lambda_c \text{CE}(\hat{c}_k, c_k)\),
where \(\lambda_u, \lambda_x, \lambda_c\) are scalar loss weights and \(\text{CE}(\cdot)\) is cross-entropy loss.
By coupling the control task with the state-prediction task, the model is encouraged to leverage the zero-torque context to refine its internal estimate of parameters \(\omega\).

\subsection{Robust Trajectory Generation via State Perturbations}
\label{gauss_per}
\rev{In closed-loop, minor prediction errors compound over time, driving the system outside the training distribution~\cite{rajaraman2020toward}.}
While methods like DAgger~\cite{ross2011reduction, celestiniTransformerBasedModelPredictive2024} address this via fine-tuning on corrected exploratory traces, we use
an alternative
requiring no additional training beyond the initial stage.
During the stabilization phase \rev{of expert rollouts,} where precise actions are key,
we use a system with a driving additive Gaussian disturbance,
\rev{
\(x_{k+1} = \Phi_\omega\big(x_k + \epsilon_k, \, \kappa_\omega(x_k + \epsilon_k),\delta_t\big)\), \(\epsilon_k \sim \mathcal{N}(0, \sigma^2 I),
\)}
forcing 
\(\kappa_\omega\) to generate corrective torques 
\rev{near the stabilization point}.
\rev{Training on these
traces equips \(\pi_\gamma\) with closed-loop recovery behaviors that prevent error compounding during inference
and mitigates numerical issues from vanishing signal magnitudes near the target.}




\section{Experimental Evaluation}
\label{experimental}
We evaluate the proposed \rev{in-context control framework on the underactuated Cartpole and Acrobot benchmark systems.}
\subsection{Model Architecture, Training and Inference Protocol}
\label{training_regime}
Each model consists of 12 layers, 8 attention heads, a 256-dimensional embedding space, and RoPE. 
\rev{\textit{Training:} We optimize \eqref{eq:loss_eqn} using AdamW with learning rate $10^{-4}$, weight decay $10^{-2}$, and batch size 64 of sub-trajectories of length $L{=}120$ from $\mathcal{T}$ as detailed in Section~\ref{univ_ctrl}. }
\textit{Inference:} At the beginning of 
inference, the model 
initially receives context \rev{$H_{0:k-1} {=}\{(x_i, e_i, c_i, u_i{=}0)\}_{i=0}^{k-1}$} 
and query \(x_k\) as a prompt and predicts $(\hat{x}_{k+1}, \hat{c}_k, \hat{u}_k)$. 
\rev{The predicted action $\hat{u}_k$ is executed in the dynamics to yield $x_{k+1}$, and the context is updated 
as $H_{1:k}{=}\{(x_i, e_i, c_i, u_i)\}_{i=1}^{k}$ 
by appending $(x_k, e_k, \hat{c}_k, \hat{u}_k)$ and dropping the oldest entry via a sliding window,}
as shown in Figure \ref{fig:inference_fig}. 
\rev{This process is repeated with context $H_{1:k}$ and query $x_{k+1}$ as the new prompt. These steps are followed until the trajectory reaches a desired length.}

\subsection{LSTM Baseline}
\label{rnn_baseline}
\comm{To test whether stabilization success is specific to ICL capabilities of Transformers
rather than a generic benefit of any sequence model, we train an
LSTM baseline (3 layers, hidden size 1024) with identical multi-head structure, training data, and loss function.
For a fair comparison, its hidden state is constructed from the same sliding context window outlined in Section~\ref{training_regime}.}

\subsection{Cartpole System}
\label{cartpole_sys}
The cartpole system requires swinging up a pendulum and balancing it upright via horizontal cart forces~\cite{aastrom2021feedback}.

\subsubsection{System Dynamics}
\label{cartpole_dynamics}
The state 
\rev{is} \(x{=}[p, \dot{p}, \theta,\dot{\theta}]^\top\), where \(p\) is cart position, \(\theta\) is the pole angle, and 
the upright equilibrium is \(x{=}[0,0,0,0]^\top\). The dynamical system is governed by 
\rev{physical parameters} \(\omega{=}[M,m,l]\in\Omega_{cart}\), where \(M\) is cart mass, \(m\) is pole mass, and \(l\) is pole length:
\begin{equation}
\begin{aligned}
    \ddot{p}&=\frac{(u+ml\dot{\theta}^2\sin{\theta})-(ml\ddot{\theta}\cos{\theta})}{M+m}\\
    \ddot{\theta}&=\frac{g\sin{\theta}+\cos{\theta}\left(\frac{-u-ml\dot{\theta}^2\sin{\theta}}{M+m}\right)}{l\left(\frac{4}{3}-\frac{m\cos^2\theta}{M+m}\right)}.
\end{aligned}
\end{equation}

\subsubsection{Parameterized Expert Controller (\(\kappa_{\omega_i}\))}
\label{ctrller_cartpole}
For each \(\omega_i\), we synthesize a hybrid expert controller \(\kappa_{\omega_i}\): \rev{(1) \emph{Energy-Shaping Swing-up:} pumps mechanical energy \(E(x;\omega_i){=}\frac{1}{2}ml^2\dot{\theta}^2{+}mgl\cos{(\theta)}\) toward upright target \(E^d{=}mgl\), applying torque \(u^\ast {=} {-}k\dot{\theta}(E {-} E^d)\) (\(k {>} 0\))~\cite{underactuated}. (2) \emph{Linear Quadratic Regulator (LQR):} near upright, switches to LQR torque \(u^\ast {=} -K(\omega_i)x\), where gain matrix \(K(\omega_i)\) solves the Algebraic Riccati Equation for linearized dynamics.}

\subsubsection{Data Generation}
\label{data_gen_cartpole}
Trajectories \(\tau^{(i)}_{\kappa_{\omega_i}} \in \mathcal{T}\) \rev{are rolled out using parameter}
realizations \(\omega_i\) from uniform distribution \(\mathcal{U}(\Omega_{\text{cart}})\), where \(\Omega_{cart}\) is defined  
in Table~\ref{tab:cartpole_params}.
Initial conditions \rev{sample} \(p_0{=}0\), \(\dot{p}_0{=}0\), \(\theta_0 {\sim} \mathcal{U}(\frac{\pi}{2},\frac{3\pi}{2})\), and \(\dot{\theta}{\sim}\mathcal{U}(-1,1)\). As stated in Section \ref{univ_ctrl}, trajectories \(\tau^{(i)}_{\kappa_{\omega_i}}\) are constructed by enforcing \(N{=}50\) steps of zero-torque evolution followed by the expert's response, and \(\delta_t=0.025\). 
\rev{To handle angular periodicity, states are augmented} as \(\tilde{x}=[p,\dot{p}, \cos(\theta), \sin(\theta),\dot{\theta}]\). Lastly, we leverage the Gaussian Perturbation approach during the LQR regime as detailed in Section \ref{gauss_per}, and failed stabilization traces are filtered out.

\begin{table}[t]
\vspace{0.5em} 
\centering
\caption{Cartpole System Parameter Space Distributions}
\label{tab:cartpole_params}
\begin{tabular}{lccc}
\toprule
\textbf{Distribution} & \textbf{$M$ (kg} & \textbf{$m$ (kg)} & \textbf{$l$ (m)} \\
\midrule
In-Distribution ($\Omega_{\text{cart}}$) & $[2.0, 3.0)$ & $[1.1, 2.0)$ & $[1.6, 2.1)$ \\
OOD Easy ($\Omega_{\text{cart}}^{\text{OOD, easy}}$) & $[1.5, 2.0)$ & $[0.3, 0.8)$ & $[1.0, 1.5)$ \\
OOD Hard ($\Omega_{\text{cart}}^{\text{OOD, hard}}$) & $[3.0, 3.5)$ & $[2.0, 2.5)$ & $[2.2, 2.7)$ \\
\bottomrule
\end{tabular}
\end{table}

\subsubsection{Main Results}
\label{results_cartpole}

\begin{figure*}[t]
    \centering
    \includegraphics[width=0.8\linewidth]{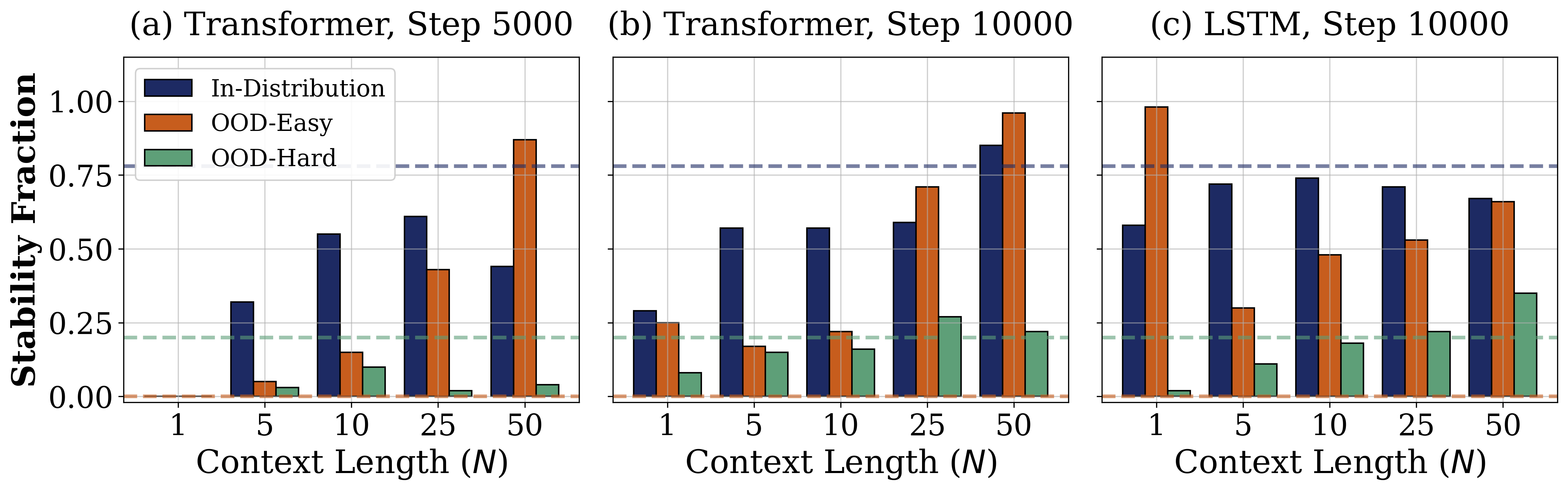}
    \caption{Stabilization success rate vs. context length ($N$) for Cartpole. Left/center: Transformer at an early and late  training checkpoint. Right: LSTM baseline at the matched late checkpoint. Dashed lines denote nominal baseline performance.
    The Transformer's ability to better exploit long context emerges between checkpoints (a)-(b), while the LSTM (c) outperforms the late-checkpoint Transformer (b) at short context but is overtaken by $N{=}50$ on in-distribution and OOD-Easy systems.}
    \label{fig:stability_cartpole}
\end{figure*}


We illustrate stabilization performance at an early \rev{(5k steps, 320k training traces)} and a late \rev{(10k steps, 640k training traces) Transformer checkpoint} 
evaluated on 100 randomly sampled systems from three distributions (Table~\ref{tab:cartpole_params}). \rev{OOD-Easy has lower inertia (lighter, shorter), making it easier to stabilize than OOD-Hard (higher inertia).}
\textit{Stability criterion:} 
\(|\theta| {\leq} 0.1 \text{ rad}\) and \(|\dot{\theta}| {\leq} 0.3 \text{ rad/s}\) 
during the final segment of the roll-out ($p$ converges slower due to underactuation).
We compare against a nominal baseline \(\kappa_{\omega_{\text{avg}}}\) using average in-distribution parameters \(\omega_{\text{avg}} = [2.5, 1.5, 1.8]\) and report the fraction of systems stabilized.

\rev{Comparing Fig.~\ref{fig:stability_cartpole}(a) and (b), in-distribution and OOD-Easy success generally increased with context until context 25 with the late-checkpoint Transformer outperforming at context 50. We also see that OOD-Easy performance exceeds in-distribution performance at longer context at both Transformer training points since training on higher-inertia systems transfers to easier, lower-inertia systems.} 
\rev{At context 1 for all data regimes, the early checkpoint stabilizes no systems, while the late checkpoint achieves non-zero success. Since one data point cannot identify three unknowns (\(\omega=[M,m,l]\)), this indicates the late checkpoint has memorized a default behavior rather than performing identification. The late checkpoint also outperforms the nominal baseline (78\% in-distribution, 0\% OOD-Easy, 20\% OOD-Hard) and is the only checkpoint able to meaningfully stabilize OOD-Hard systems, suggesting this capability likewise benefits from extended training.}
This needs to be explored in future work.
\comm{For OOD-Hard systems specifically, success degrades at the longest context (50), consistent with ICL literature showing that additional context can introduce spurious correlations that mislead the model~\cite{chen2023many}, an effect likely amplified since OOD-Hard 
systems are much harder to stabilize.}


\comm{We compare the late-checkpoint Transformer 
against an LSTM baseline (Fig.~\ref{fig:stability_cartpole}(c)) trained identically.
At context 1, the LSTM substantially outperforms the Transformer for in-distribution and OOD-Easy systems, reflecting a stronger memorized default policy rather than superior context utilization. This advantage narrows and reverses with context, and by context 50, the Transformer outperforms the LSTM on both in-distribution and OOD-Easy systems, while the LSTM retains better performance on OOD-Hard systems, where identification is most difficult. }

\begin{figure}[t]
    \centering
    \includegraphics[width=0.75\linewidth]{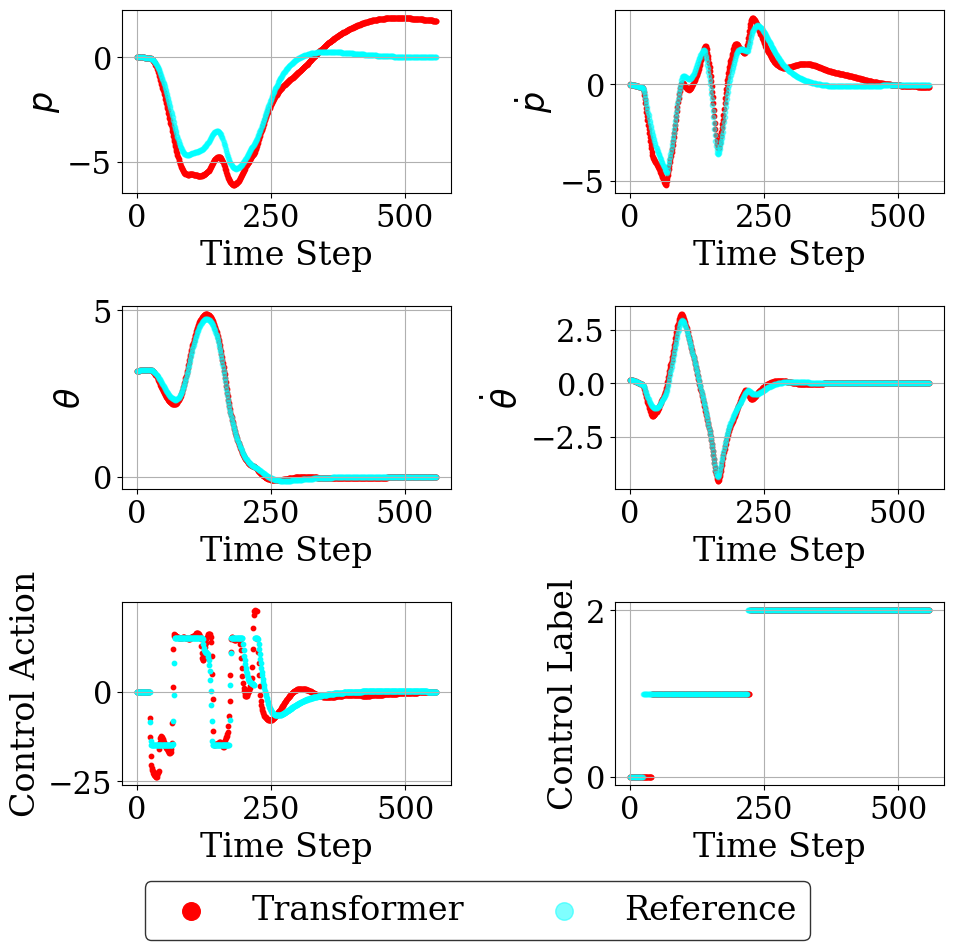}
    \caption{One example of swing-up for the cartpole system. Comparison between expert reference \(\kappa_\omega\) and Transformer roll out given \(25\) context examples. States \([p,\dot{p},\theta,\dot{\theta}]\) and actions are shown over time.}
    \label{fig:trajectories_cartpole}
\end{figure}

\rev{Fig.~\ref{fig:trajectories_cartpole} shows one example of a Transformer rollout under \(N{=}25\) context steps versus expert reference \(\kappa_\omega\).}
While the predicted Transformer actions are not perfectly aligned with \(\kappa_\omega\), $\theta$ and $\dot{\theta}$
converge to zero indicating stabilization.
Notably, the Transformer nearly perfectly predicts the transition from energy-shaping swing-up 
to LQR stabilization.

\subsubsection{Ablation}
\label{ablation_noise}

\begin{table}[t!]
\centering
\caption{Stabilization Success vs. Context Length \(N\) for the Cartpole Transformer (Ablation: Noise-Free Data)}
\label{tab:context_results}
\begin{tabular}{@{}lccccc@{}}
\toprule
\textbf{Test Distribution} & \textbf{N=1} & \textbf{N=5} & \textbf{N=10} & \textbf{N=25} & \textbf{N=50} \\
\midrule
In-Distribution   & 0.00 & 0.02 & 0.02 & 0.01 & 0.05 \\
Out-of-Distribution (OOD) & 0.00 & 0.01 & 0.03 & 0.04 & 0.09 \\
\bottomrule
\end{tabular}
\end{table}

A key hypothesis in our framework, as discussed in Section~\ref{gauss_per}, is that training on traces synthesized by expert controllers \(\kappa_{\omega_i}\) injected with additive Gaussian noise during the LQR phase encourages the Transformer to learn corrective actions necessary to maintain stabilization. To validate this claim, \rev{we train a Transformer 
on ``clean'' dataset \(\mathcal{T}_{\text{clean}}\) lacking Gaussian noise.}
As shown in Table~\ref{tab:context_results}, the noise-free model
fails \rev{on $>90\%$ of test systems. Without recovery demonstrations near equilibrium, minor inference errors drive states out-of-distribution, proving expert stochasticity is essential for closed-loop robustness.}


\subsection{Acrobot System}
\label{acrobot_sys}
The Acrobot is a two-link planar robot actuated only at the elbow joint, which is a significantly more challenging benchmark to swing up and balance ~\cite{underactuated}.

\subsubsection{System Dynamics}
The state is 
\(x{=}[\theta_1, \theta_2, \dot{\theta}_1,\dot{\theta}_2]^\top\), where \(\theta_1\) first link angle, \(\theta_2\) is the second link angle relative to the first, and 
the upright equilibrium is \(x{=}[\pi,0,0,0]^\top\). 
Physical parameter vector $\omega=[m_1,m_2,l_1,l_2]$ denotes link masses $m_1, m_2$ and link lengths $l_1, l_2$. The equations of motion are expressed as:
\begin{equation}
    \mathbf{M}(\mathbf{q},\omega)\ddot{\mathbf{{q}}}+\mathbf{C}(\mathbf{q},\dot{\mathbf{q}},\omega)\dot{\mathbf{q}}=\mathbf{G}(\mathbf{q},\omega) +\mathbf{B}u,
\end{equation}
where 
\(\mathbf{q}=[\theta_1,\theta_2]^\top\) and matrices $\mathbf{M}, \mathbf{C}, \mathbf{G}$ follow
standard manipulator notation~\cite{underactuated}.
\subsubsection{Parameterized Expert Controller (\(\kappa_{\omega_i}\))}
For each realization \(\omega_i\), we synthesize a hybrid expert controller \(\kappa_{\omega_i}\)  with hysteresis switching logic to prevent chattering: (1) \textit{Partial Feedback Linearization Swing-up:} \rev{\(u=\bar{d}_{22}v_2+\bar{h}_2+\bar{\phi}_2\) where \(\bar{d}_{22},\bar{h}_2,\bar{\phi}_2\) are reduced-order terms derived from the dynamics, and \(v_2 = \text{taper}(\theta_1^d,\theta_1) \left[ k_p(E) (\theta_{2}^d - \theta_2) - k_d(E) \dot{\theta}_2 \right]\) is an energy-shaping signal~\cite{spong2002swing}. (2) LQR: near upright, switches to LQR torque $u^\ast {=} -K(\omega_i)(x - x_{\text{goal}})$.}

\subsubsection{Data Generation}
\label{data_gen_acrobot}
\rev{Trajectories $\tau^{(i)}_{\kappa_{\omega_i}} \in \mathcal{T}$ are rolled out using sample realizations $\omega_i \sim \mathcal{U}(\Omega_{\text{acro}})$ bounded as in Table~\ref{tab:acrobot_params}. Initial conditions sample $\theta_1 \in \mathcal{U}(-\frac{\pi}{2},-\frac{\pi}{4}) \cup \mathcal{U}(\frac{\pi}{4},\frac{\pi}{2})$, $\theta_2=0$, $\dot{\theta}_1 \sim \mathcal{U}(-1,1)$\,rad/s, and $\dot{\theta}_2=0$. Trajectories $\tau^{(i)}_{\kappa_{\omega_i}}$ are constructed by enforcing $N{=}100$ steps of zero-torque evolution preceding the expert's ($\kappa_{\omega_i}$) response, and $\delta_t{=}0.02$. To accommodate angular periodicity, states are augmented as}
\(x=[\cos(\theta_1),\sin(\theta_1),\cos(\theta_2),\sin(\theta_2),\dot{\theta}_1,\dot{\theta}_2]\).
We also apply Gaussian perturbation during the LQR regime addressing concerns mentioned in Section \ref{gauss_per}. 
Unstabilized traces are filtered out ($\sim 33\%$ of trajectories due to system complexity).

\begin{table}[t]
\centering
\caption{Acrobot System Parameter Space Distributions}
\label{tab:acrobot_params}
\begin{tabular}{lcccc}
\toprule
\textbf{Distribution} & \textbf{$l_1$ (m)} & \textbf{$l_2$ (m)} & \textbf{$m_1$ (kg)} & \textbf{$m_2$ (kg)} \\
\midrule
In-Distr. ($\Omega_{\text{acro}}$) & $[0.5, 1.0)$ & $[1.0, 1.5)$ & $[1.5, 2.0)$ & $[1.0, 1.5)$ \\
OOD ($\Omega_{\text{acro}}^{\text{OOD}}$) & $[1.0, 1.5)$ & $[1.5, 2.0)$ & $[1.0, 1.5)$ & $[1.0, 1.5)$ \\
\bottomrule
\end{tabular}
\end{table}

\subsubsection{Main Results}
\label{results_acrobot}
\begin{figure*}[t!]
    \centering
    \includegraphics[width=0.8\linewidth]{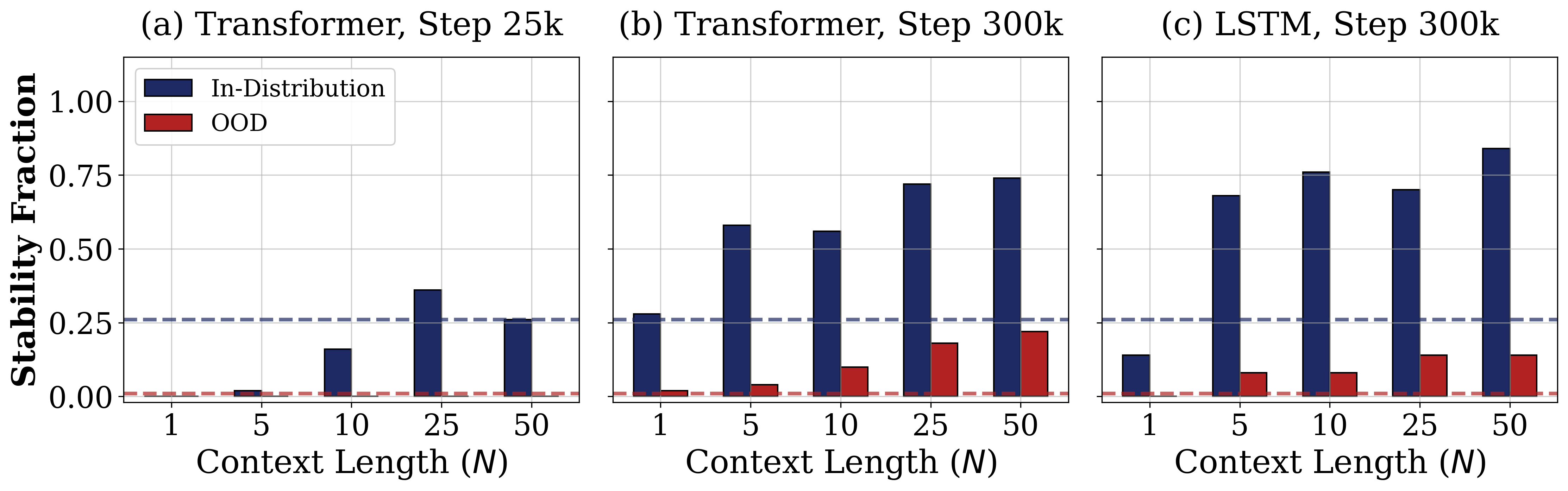}
        \caption{Stabilization success rate vs.\ context length ($N$) for Acrobot. (a) Transformer, early checkpoint (step 25k). (b) Transformer, late checkpoint (step 300k). (c) LSTM baseline, matched late checkpoint (step 300k). Dashed lines denote nominal baseline performance. The LSTM matches or exceeds the Transformer on in-distribution systems at all context lengths beyond $N{=}1$, while the Transformer overtakes the LSTM on OOD systems by $N{=}50$.}
    \label{fig:stability_acrobot}
\end{figure*}

\begin{figure}[t!]
    \centering
    \includegraphics[width=0.75\linewidth]{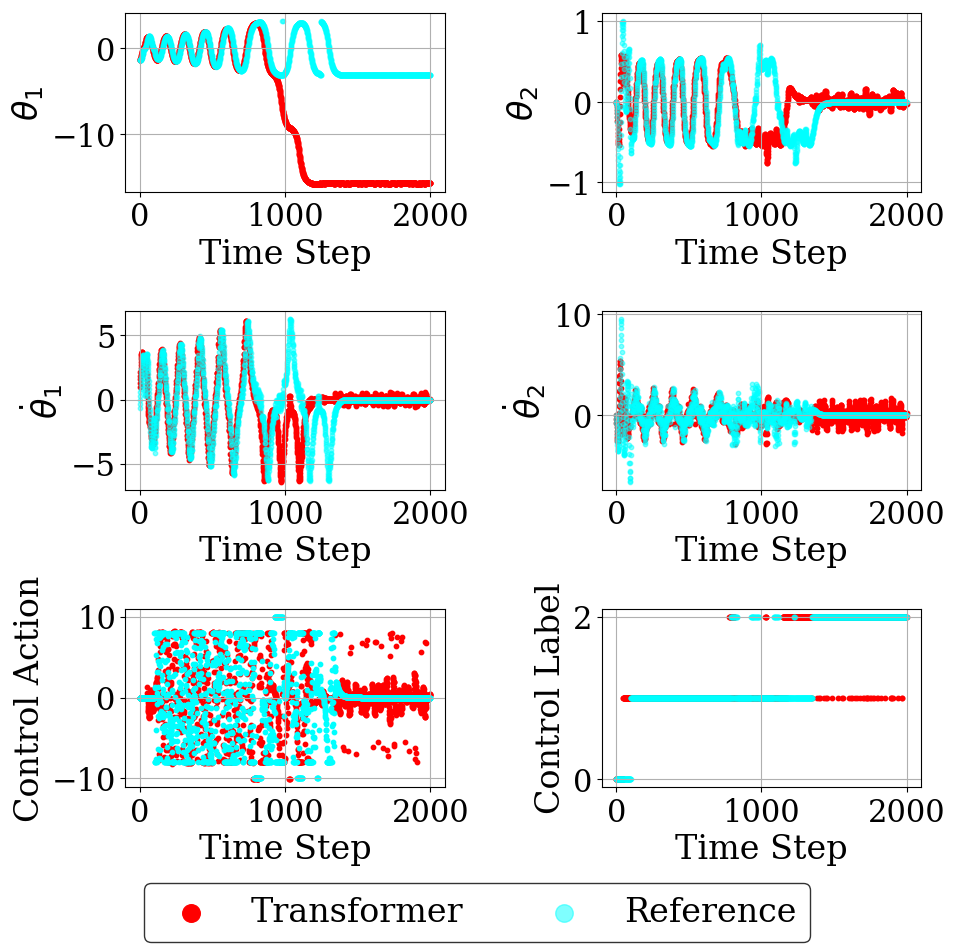}
        \caption{One example of acrobot swing-up rollout comparison: Expert $\kappa_\omega$ (light blue) vs. Transformer (red) given $N{=}50$ context steps. Angle $\theta_1$ converges to $-5\pi \equiv \pi$ rad upright.}
        \label{fig:trajectories_acrobot}
\end{figure}

We evaluate \rev{performance} on 50 in-distribution
and 50 OOD 
randomly sampled test systems (Table~\ref{tab:acrobot_params}). 
\textit{Stability Criterion:}
given the complexity of the acrobot, \rev{a system is stabilized if the state enters and remains within a fixed neighborhood (\(|\theta_1|<0.2 \text{ rad and } |\theta_2|<0.3 \text{ rad}\)) of upright equilibrium for at least 3 seconds,}
indicating that the model can swing up and switch to LQR stabilization, even briefly.
\rev{We compare against a nominal baseline $\kappa_{\omega_{\text{avg}}}$ using average in-distribution parameters $\omega_{\text{avg}} = [l_1=0.75, l_2=1.25, m_1=1.75, m_2=1.25]$, which stabilizes only 26\% of in-distribution and 1\% of OOD systems.}

\rev{Fig.~\ref{fig:stability_acrobot}(a) indicates that the early-checkpoint Transformer (25k training steps, 1.6 million training trajectories) stabilizes no systems at \(N{=}1\) but improves up to context 25, confirming context-driven adaptation rather than memorization. The late checkpoint (300k training steps, 20.5 million training traces) shown in Fig.~\ref{fig:stability_acrobot}(b) achieves strong in-distribution stabilization and increasingly leverages context for OOD systems,
exceeding the nominal baseline. Non-zero stabilization success for context 1 indicates memorization which
seems to help with OOD stabilization.}

\comm{Comparing the late checkpoint Transformer 
with the LSTM in Fig.~\ref{fig:stability_acrobot}(c), the LSTM matches or exceeds the Transformer's in-distribution performance at every context length beyond \(N{=}1\). 
So, the in-context learning abilities of Transformer models do not outperform LSTM behavior in these examples.
On OOD systems, however, the pattern resembles the Cartpole for OOD-Easy: the LSTM has better performance at short context, but the Transformer does better at longer context \(N{=}50\) (0.22 vs.\ 0.14). Taken together with the Cartpole OOD-Hard result, where the LSTM retained an advantage even at long context, this suggests that the Transformer's context driven advantage over the LSTM is not uniform, but perhaps Transformers have an advantage in performing in unfamiliar regimes. }

Fig.~\ref{fig:trajectories_acrobot} shows one example of a Transformer rollout under $N{=}50$ context examples versus reference expert controller \(\kappa_\omega\). \rev{The Transformer successfully completes the swing-up maneuver, converging $\theta_1 \to -5\pi \equiv \pi$\,rad and applies LQR corrective actions to maintain upright stability.}

\comm{\textit{Limitations:} Our framework requires an expert controller \(\kappa_\omega\) per training realization. This is tractable for Cartpole and Acrobot since swing-up/LQR expert is parameterized by \(\omega\) but may not generalize to systems lacking an analytic expert.}

\section{Conclusion}
We investigated the ICL capabilities of Transformer models for the low-level stabilizing control of underactuated, nonlinear dynamical systems. Provided with a context of zero-torque state-action observations, the Transformer can adapt its control law accordingly. We find that it is not possible to successfully train on data from systems with no driving disturbance. However, training on trajectories from systems with additive Gaussian noise 
allows the model to learn the corrective actions required for stabilization that are otherwise absent in ``clean'' expert trajectories.
Furthermore, we show that once the Transformer begins to memorize the training data and stabilize systems with no context examples, ICL capabilities seem to emerge for even for OOD systems. Evaluated on cartpole and acrobot benchmarks, the Transformer consistently outperforms naive nominal baselines but has similar performance to an LSTM baseline. 

\bibliography{reference}
\bibliographystyle{ieeetr}
\end{document}